\documentclass{article}
\usepackage{ijcai26}

\usepackage{times}
\usepackage{soul}
\usepackage{url}
\usepackage[hidelinks]{hyperref}
\usepackage[utf8]{inputenc}
\usepackage[small]{caption}
\usepackage{graphicx}
\usepackage{amsmath}
\usepackage{amsthm}
\usepackage{booktabs}
\usepackage{multirow}
\usepackage[switch]{lineno}

\usepackage{algorithmicx}
\usepackage{algpseudocode}
\usepackage[linesnumbered,ruled,vlined]{algorithm2e}
\usepackage{amssymb}
\usepackage{xcolor}

\title{CoDG-Net: Structure-Guided Style Diffusion and Collaborative Learning to Mitigate Catastrophic Forgetting in Medical Image Domain Generalization}

\author{
Yucheng Song$^1$
\and
Jincan Wang$^2$\and
Haokang Ding$^1$\and
Zhiqiang Tian$^2$ \and
Kangxu Fan$^3$ \And
Zhifang Liao$^{1,*}$
\\
\affiliations
$^1$School of Computer Science and Engineering, Central South University, Changsha, China\\
$^2$School of Software Engineering, Xi’an Jiaotong University, Xi’an, Shaanxi, China\\
$^3$College of Computer Science and Artificial Intelligence, Fudan University, Shanghai, China\\
}

\begin{document}

\maketitle

\begin{abstract}
Domain Generalization (DG) for medical image segmentation is both highly challenging and critically important. However, existing medical DG methods largely overlook the issue of Catastrophic Forgetting (CF): \textbf{Models often sacrifice their ability to retain source-domain knowledge while pursuing cross-domain robustness.} This can directly threaten diagnostic safety in already-deployed clinical scenarios. To address this, we investigate data augmentation strategies and catastrophic forgetting for medical image DG segmentation. First, we propose a structure-guided style diffusion augmentation method. Constrained by anatomical structure consistency in the frequency domain, this method performs cross-domain diffusion on the amplitude spectrum, generating samples with more diverse and broader style coverage to better support domain generalization. Then, we design a collaborative learning network with a dual-branch interactive architecture (CoDG-Net), together with a novel learning bias-guided strategy that adaptively regulates knowledge transfer at both the layer level and the task level, thereby effectively mitigating catastrophic forgetting on the source domain. Experiments and ablation studies on single-source and multi-source medical DG benchmark datasets demonstrate that CoDG-Net not only outperforms existing state-of-the-art methods in target-domain segmentation performance, but also achieves a lower forgetting rate on the source-domain data. The code is available at: \url{https://github.com/wangprocess/CoDG-Net}.
\end{abstract}

\section{Introduction}
Domain Generalization (DG) in medical imaging is a key technique for achieving robust medical image segmentation under multi-center and multi-device conditions \cite{li2020domain,li2021progressive,li2021medical,wang2022generalizing}. DG requires the model to rely solely on source-domain samples and maintain high performance on unseen target domains, without accessing any target-domain data during training \cite{su2023rethinking,li2024intra,cheng2023adversarial,robey2021model,liu2021feddg,robey2021model}. In medical imaging, this capability is not only about cross-domain performance, but is also directly related to the accuracy and safety of clinical diagnosis \cite{liu2025perturbating}. However, in medical image DG segmentation, the problem of Catastrophic Forgetting (CF) has long been overlooked. Due to privacy and regulatory constraints, source-domain data cannot be freely transferred across institutions, making common continual learning strategies such as mixed-data training or replay of past data infeasible in medical scenarios. This makes CF even more challenging in medical image DG tasks.
\begin{figure}[ht]
  \centering
  \includegraphics[width=\linewidth]{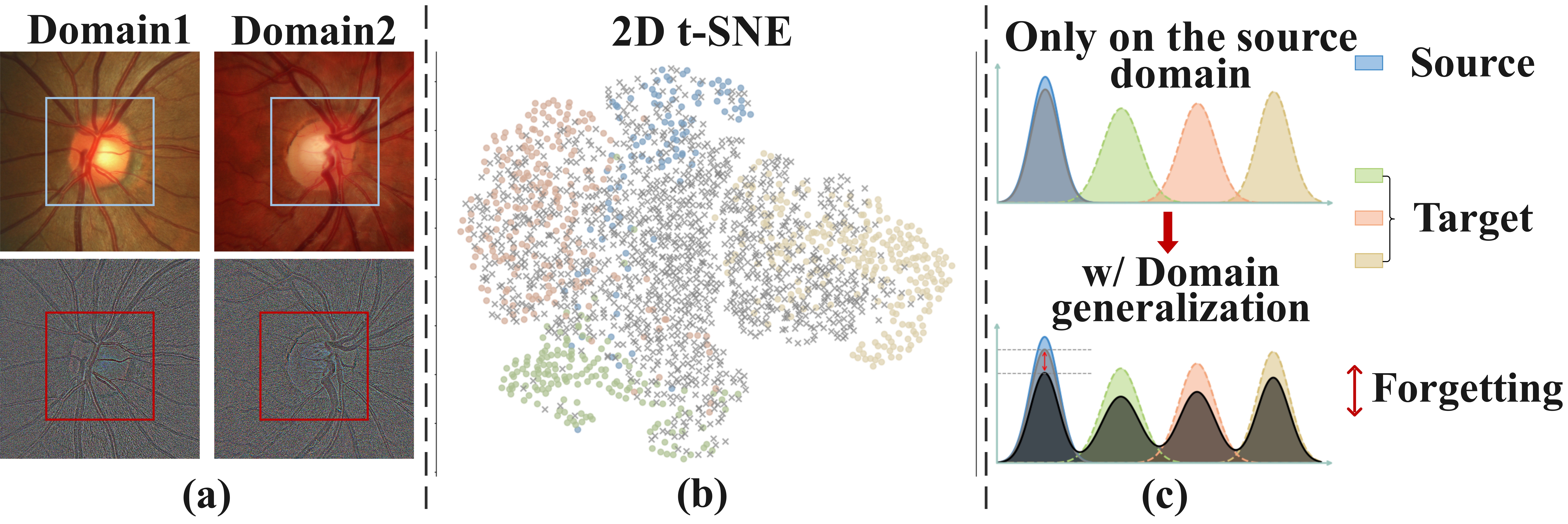}
  \caption{Conceptual overview of our research motivation. (a) Original amplitude and phase spectra of fundus images acquired from different devices. (b) Coverage of extended domains after applying traditional data augmentation and diffusion-based generation methods on source domain 1. (c) Comparison of distribution-fitting capability on the source domain.}
  \label{intro}
\end{figure}

To simultaneously improve model generalization and mitigate the CF problem, we conduct a statistical analysis of feature distributions on cross-domain data generated by augmentation methods and derive our motivations as follows: (1) Cross-domain data exhibit consistent anatomical structures, while their differences mainly lie at the style level, such as contrast, texture, and brightness (i.e., style inconsistency). For example, as shown in Fig.\ref{intro}(a), fundus images acquired from different devices share similar anatomical structures, whereas inter-domain discrepancies are primarily reflected in contrast and brightness; (2) Combining traditional augmentations with diffusion-based generation can better cover the cross-domain data distribution and thus enhance domain robustness. As illustrated in Fig.\ref{intro}(b), applying traditional augmentation together with diffusion-based generation on the source domain yields extended domains with richer style information; (3) Unlike training that only fits the source-domain distribution, DG imposes additional generalization constraints (e.g., style diversity) on the parameters within a single training phase to improve cross-domain robustness. This drives the model away from the “source-domain optimum” toward a compromise that is more robust but no longer optimal on the source domain, thereby reducing its distribution-fitting capability on the source domain, as shown in Fig.\ref{intro}(c).

Motivated by the above, we aim to simultaneously improve the model’s generalization ability on unseen domains and effectively alleviate catastrophic forgetting. First, we propose a novel structure-guided style diffusion augmentation method. It uses the Fourier transform to extract frequency-domain signals from pseudo-domain images generated by traditional style augmentation methods. We then perform diffusion generation on the amplitude spectrum (style), and recombine it with the phase spectrum to obtain new samples with a broader coverage of feature distributions. To effectively mitigate catastrophic forgetting, we design a collaborative learning network with a dual-branch interactive structure (CoDG-Net). The network contains two complementary branches: a source-domain image restoration branch and a generalized segmentation branch. To promote sufficient interactive optimization between the feature spaces of these two branches and to balance source-domain knowledge preservation with target-domain generalization ability, we propose a novel learning bias–guided strategy. It adaptively regulates knowledge transfer at both the layer level and the task level, improving segmentation accuracy while effectively alleviating the CF problem. The main contributions of this paper are summarized as follows:
\begin{itemize}
    \item We propose a structure-guided style diffusion augmentation method. Under the constraint of preserving anatomical structure consistency, this method introduces a diffusion model to better fit the style distributions of unknown domains, while performing diffusion only on the amplitude spectrum to significantly reduce the difficulty of generation.
    \item We investigate the CF problem in medical image domain generalization. We propose a collaborative learning network with a dual-branch interaction structure and design a novel learning bias-guided strategy for it, which guides knowledge transfer in the network and effectively alleviates this problem.
    \item Experimental results show that our method outperforms existing SOTA methods on both single-source and multi-source domain generalization datasets, and achieves a lower forgetting rate on source-domain data.
\end{itemize}

\section{Related Work}
\subsection{Domain Generalization in Medical Imaging}
Domain generalization (DG) aims to train a model on one or multiple source domains so that it can directly generalize to unseen target domains, which is crucial for robust medical image analysis under varying conditions \cite{yoon2024domain}. To enhance the cross-domain generalization ability of medical image segmentation models, some medical DG methods adopt data augmentation techniques—such as geometric/color perturbations, frequency-domain perturbations, and style transfer—to enrich style variations of source-domain samples and simulate potential target-domain distributions \cite{liu2024universal,huang2025generative,dou2019domain}. Other methods exploit domain-invariant feature extraction, cross-domain alignment, adversarial learning, or normalization strategies to reduce inter-domain discrepancies  \cite {choi2023progressive,10371364,zhou2022generalizable,seo2020learning}.


\subsection{Catastrophic Forgetting in Continual Learning}

Catastrophic Forgetting (CF) is a core problem in Continual Learning (CL), referring to the phenomenon that a model significantly forgets previously learned knowledge when learning new tasks or new sites \cite{wang2024comprehensive}. Common mitigation strategies include parameter regularization, knowledge distillation, and data replay \cite{nori2025autoencoder,zhao2024statistical,ma2022continual}. In CL-based solutions for DG, Xu et al. \cite{xu2025towards} address cross-site continual segmentation with a joint pursuit of memorization and generalization, and alleviate CF via site-modulated replay and gradient alignment. However, their method assumes that multi-site data arrive sequentially and explicitly relies on replayed samples, essentially following a replay-based strategy; moreover, historical site data remain accessible during training.

In contrast, the medical image DG setting considered in this paper has no access to target-domain data during training, thus lacking the cross-domain alignment conditions that most Domain Adaptation (DA) methods rely on. The model is trained only once on a fixed source domain, and after training it can neither access the original source data nor perform any form of task-level data replay. Therefore, existing replay-based CL methods are difficult to directly transfer to the DG setting. To address these limitations, we propose CoDG-Net, which tackles the CF problem in DG under the constraints of no replay and no target-domain data.

\section{Method}
\begin{figure*}[ht]
  \centering
  \includegraphics[width=\linewidth]{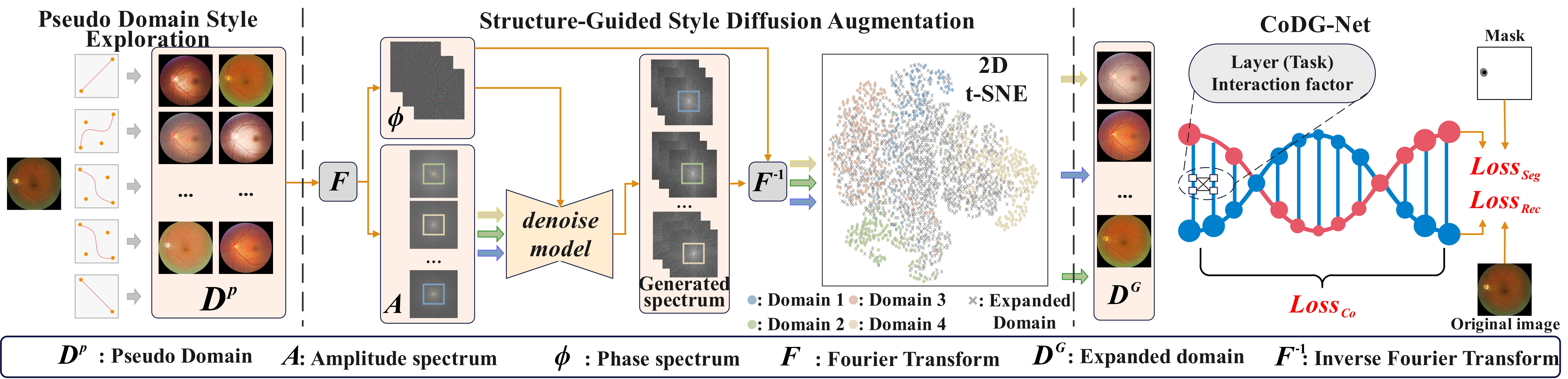}
  \caption{Overall workflow of the proposed method. First, pseudo-domain style exploration is applied to obtain \(D^P\). Second, a diffusion-based style diversification method is used to generate a style-diverse \(D^G\). Finally, the generated data are fed into the proposed CoDG-Net for training.}
  \label{method}
\end{figure*}
\subsection{Problem Definition and Overview}
In domain generalization tasks, there exist a source domain \(D^S\) and a target domain \(D^T\). The source domain is defined as $D^S = \{x_i^S, y_i^S\}_{i=1}^{N_S}$, where \(x_i^S\) is the \(i\)-th sample in the source domain, \(y_i^S\) is the corresponding segmentation mask, and \(N_S\) is the total number of samples. The goal of this paper is to train a segmentation model \(S_{\text{seg}}\) using only samples from \(D^S\), and to ensure its generalization ability to unseen target domains.

As shown in Fig. \ref{method}, to address the problem of limited style diversity in source-domain data, we first adopt a traditional Bézier-curve–based intensity mapping strategy \cite{zhou2022generalizable} to explore pseudo-domain styles and obtain a pseudo-domain \(D^P\). We then introduce a structure-guided style diffusion augmentation module. This module exploits the anatomical consistency of medical images and performs style expansion on source-domain data in the Fourier domain, generating an extended domain $D^G = \{x_i^G, y_i^G\}_{i=1}^{N_G}$ with diverse style distributions. Here, each \(x_i^G\) preserves the structural semantics of \(x_i^S\) while simulating the style characteristics of potential target domains. On this basis, in order to adapt to the distribution of \(D^G\) while suppressing catastrophic forgetting, we propose a collaborative learning network with a dual-branch interactive structure (CoDG-Net). One branch trains a segmentation model $S_{\text{seg}}: x^G \rightarrow y^G$, and the other branch trains a reconstruction model $S_{\text{rec}}: x^G \rightarrow x^S$. Through the collaborative training of these two tasks, CoDG-Net establishes a strong interaction mechanism in the feature space, effectively balancing cross-domain generalization and source-domain memory preservation.

\subsection{Structure-Guided Style Diffusion Augmentation}
In previous studies, a large body of work has demonstrated that image style and structure are encoded in the amplitude spectrum and phase spectrum of the Fourier domain, respectively \cite{qiao2024medical,lv2022causality,piotrowski1982demonstration,oppenheim2005importance}. Therefore, to further enhance sample diversity and improve the model’s coverage of potential target domains, we propose a novel structure-guided style diffusion augmentation method. The core idea of this method is to exploit the explicit constraint of image structure provided by the Fourier phase spectrum, and to combine it with the strong generative capability of diffusion models to explore unknown amplitude (style) distributions. As shown in Fig. 3, this process not only generates style-diverse \(D^G\), but also strictly preserves anatomical structural consistency, thereby avoiding structural distortions in the generated images.

\textbf{Forward Process:} First, we apply the fast Fourier transform (FFT) to convert the input pseudo-domain image into the frequency domain and decouple it into an amplitude spectrum \(A(u, v)\) and a phase spectrum \(\Phi(u, v)\). We take the amplitude spectrum as the diffusion target, while the phase spectrum is preserved for subsequent structural guidance. Specifically, in the forward noising process, we degrade the amplitude spectrum by adding Gaussian noise, as follows:

\begin{equation}
q(x_t \mid x_0) = \mathcal{N}\left(x_t; \sqrt{\bar{\alpha}_t} \cdot x_0,\; (1 - \bar{\alpha}_t) \cdot I\right)
\end{equation}

Where \(x_0\) denotes the original amplitude spectrum data, and \(x_t\) is the noisy data at time step \(t\). The term \(\bar{\alpha}_t = \prod_{s=1}^{t} \alpha_s\) is the cumulative attenuation coefficient, where \(\alpha_t = 1 - \beta_t\) and \(\beta_t\) represents the noise strength at each time step. \(\mathcal{N}(\cdot; u, \sigma^2)\) denotes a normal distribution with mean \(u\) and variance \(\sigma^2\).

\textbf{Reverse Process:} The goal of the reverse denoising process is to restore the original data \(x_0\) from the noisy data \(x_t\), by introducing \(\Phi(u, v)\) as a structural prior condition into the denoising network. The process is defined as:

\begin{equation}
    p_\theta(x_{t-1} \mid x_t, \Phi) = \mathcal{N}\bigl(x_{t-1};\, u_\theta(x_t, t, \Phi),\, \Sigma_\theta(x_t, t)\bigr)
\end{equation}

Where \(\Sigma_\theta(x_t, t)\) is the variance term, representing the uncertainty of generation; \(\theta\) denotes the learnable parameters; and \(u_\theta(x_t, t, \Phi)\) is the mean prediction function conditioned on the constraint information \(\Phi\). It represents the prediction of \(x_{t-1}\) from \(x_t\) at step \(t\), and is given by:

\begin{equation}
    u_\theta(x_t, t, \Phi) = \frac{1}{\sqrt{\alpha_t}} \left( x_t - \frac{\beta_t}{\sqrt{1 - \bar{\alpha}_t}} \, \epsilon_\theta(x_t, t, \Phi) \right)
\end{equation}

Where \(\epsilon_\theta(x_t, t, \Phi)\) is the noise component predicted by the model. The model learns to approximate the noise \(\epsilon\) added in the forward process, based on \(x_t\), the time step \(t\), and the constraint condition \(\Phi\).

\textbf{Sampling process:} We propose a continuous compositional sampling mechanism in the frequency domain, which aims to generate samples that are both style-diverse and structurally reliable. Specifically, we first apply the Fourier transform to extract the phase spectrum \(\Phi(\cdot)\) and the amplitude spectrum \(A(\cdot)\) of the original image. Then, we sample an initial noise \(x_T \sim \mathcal{N}(0, I)\) from a standard normal distribution as a latent style code. Under the guidance of the original \(\Phi(\cdot)\), we perform the reverse denoising process to generate a new amplitude spectrum \(A'(u,v)\). Finally, we recombine the generated \(A'(\cdot)\) with the original \(\Phi(\cdot)\), and reconstruct the final extended-domain image \(f(x,y)\) via the inverse Fourier transform (iFFT):

\begin{equation}
    f(x,y) = \mathcal{F}^{-1}\left(A'(u,v) \cdot e^{i\Phi(u,v)}\right)
\end{equation}

Since this method performs diffusion-based generation only on the amplitude spectrum, it substantially reduces the difficulty of the diffusion process. At the same time, the phase spectrum serves as guidance to preserve critical lesion and organ boundaries with pixel-level accuracy.

\begin{figure}[ht]
  \centering
  \includegraphics[width=\linewidth]{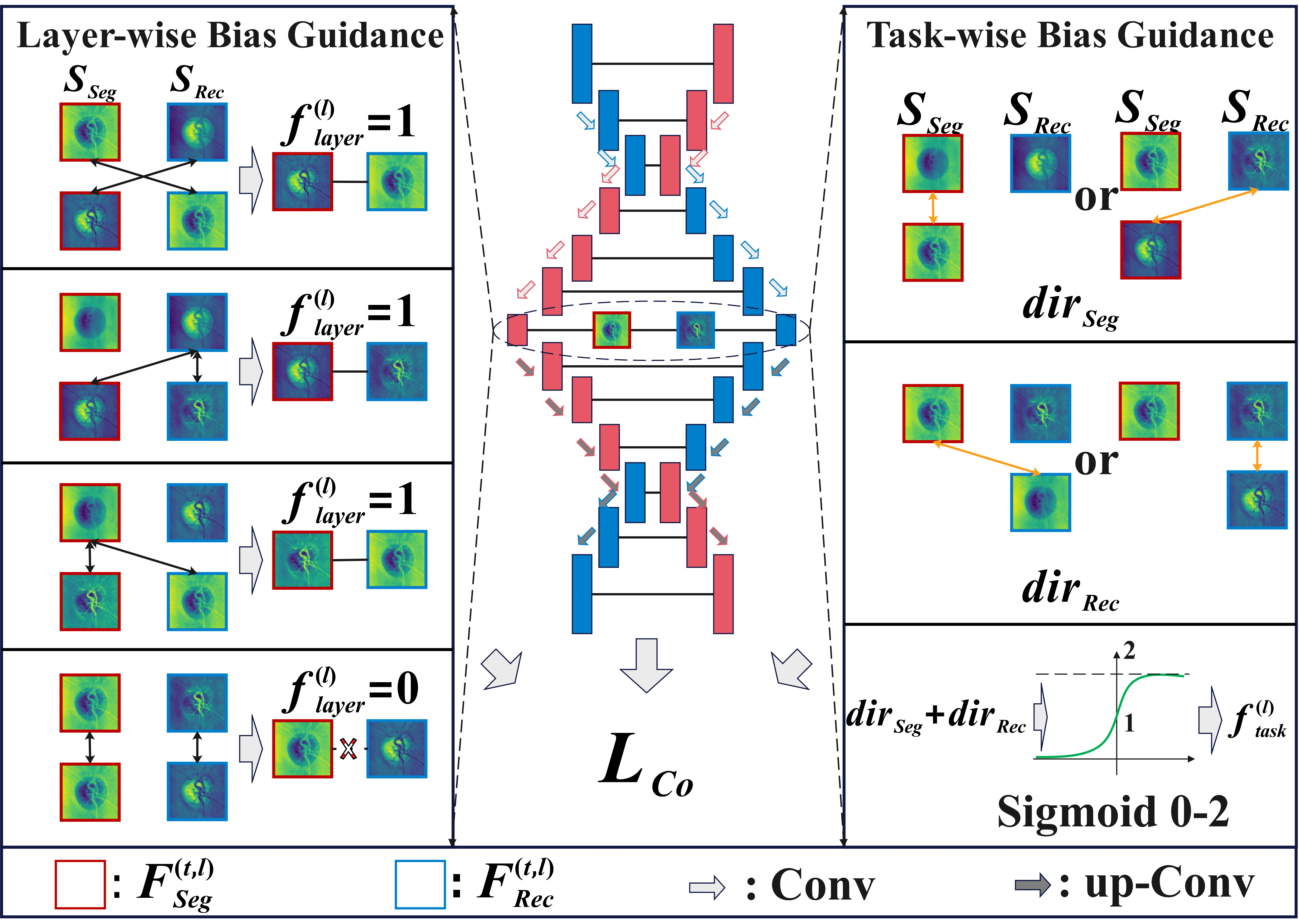}
  \caption{Overall architecture of the proposed CoDG-Net and the bias-guided learning strategies (for clarity, skip connections within each U-Net branch are omitted).}
  \label{net}
\end{figure}

\subsection{Collaborative Learning Network with Dual-Branch Interaction}
To explicitly model the cross-domain style mapping patterns, we design CoDG-Net with two parallel branches, \(S_{\text{rec}}\) and \(S_{\text{seg}}\) (as shown in Fig. \ref{net}). The network captures style mapping information indirectly through collaborative learning. Specifically, \(S_{\text{rec}}\) is dedicated to learning the inverse mapping “from the extended domain back to the source domain”: when the network attempts to restore multi-style extended samples to the source-domain appearance, the intermediate layers inevitably encode rich style transfer relationships. These inter-domain mapping cues are then passed to \(S_{\text{seg}}\) via feature interaction between the two branches, enabling \(S_{\text{seg}}\) to truly learn cross-domain appearance variations during training, rather than merely adding generalization regularization around the source-domain optimum. 

Moreover, not all intermediate-layer features are mutually beneficial for the two branches. If cross-branch interactions are indiscriminately enabled at every layer, or if a unified soft-gating/attention fusion is applied across all layers, task-irrelevant noise can be propagated and the reconstruction branch may exert excessive interference on the main segmentation branch, thereby degrading segmentation performance. Therefore, we propose a Learning Bias-Guidance Strategy. The layer-wise strategy selectively opens feature interactions at key layers to suppress ineffective information propagation, while the task-wise strategy explicitly biases the global optimization toward the segmentation task, thus enhancing cross-domain robustness and effectively mitigating, or even reversing, catastrophic forgetting on the source domain.

\subsubsection{Layer-Wise Bias-Guided Strategy}
In multi-task collaborative learning, feature layers at different network depths usually carry different semantic or style information, and not all layers need cross-task interaction at all times. To adaptively control the interaction behavior of each layer, we propose a layer-wise bias guidance strategy. This strategy aims to decide whether the current layer should focus on “self feature enhancement” or “cross-task feature fusion”. Specifically, we define an Interaction Flag to replace traditional hard-coded connections. For the \(l\)-th layer of the network, between training steps \(t\) and \(t + k\), we compute the evolution trend of the feature map \(F^{(t, l)}\). We use the Euclidean distance \(d(\cdot, \cdot)\) to measure the consistency change of features. If the feature \(F^{(t, l)}\) of one branch at step \(t\) is closer to the feature \(F_{\text{other}}^{(t+k, l)}\) of the other branch after \(k\) update steps, it indicates that this layer tends to absorb knowledge from the other task, and we mark it as a Collaborative State. Conversely, if the feature is more inclined to maintain its own update direction, we mark it as an Independent State. To prevent the model from falling into a local optimum or staying in a single state for a long time, we introduce a stochastic perturbation mechanism, which forcibly flips the current state with probability \(\delta\). Finally, according to the computed state, we generate a layer-wise interaction factor \(f_{\text{layer}}^{(l)} \in \{0, 1\}\) to dynamically turn on or off the feature interaction channel at that layer (see Algorithm \ref{alg1} for the detailed procedure).

\begin{algorithm}
    \caption{Layer-wise Bias-Guided Strategy.}
    \label{alg1}
    \KwIn{training step \(t\); intermediate feature maps \(F_{\text{seg}}^{(t,l)}\) and \(F_{\text{rec}}^{(t,l)}\); feature maps after \(k\) update steps \(F_{\text{seg}}^{(t+k,l)}\) and \(F_{\text{rec}}^{(t+k,l)}\)}

    $\text{Flag}_{\text{seg}} \leftarrow d\bigl(F_{\text{seg}}^{(t,l)}, F_{\text{seg}}^{(t+k,l)}\bigr) > d\bigl(F_{\text{seg}}^{(t,l)}, F_{\text{rec}}^{(t+k,l)}\bigr)$

    $\text{Flag}_{\text{rec}} \leftarrow d\bigl(F_{\text{rec}}^{(t,l)}, F_{\text{rec}}^{(t+k,l)}\bigr) > d\bigl(F_{\text{rec}}^{(t,l)}, F_{\text{seg}}^{(t+k,l)}\bigr)$

    \algorithmiccomment{True → collaborate; False → independent}

    With probability \(\delta\): $\text{Flag} \leftarrow \neg \text{Flag}$

    Compute state: $\text{State} \leftarrow \text{Flag}_{\text{seg}} \lor \text{Flag}_{\text{rec}}$

   \If{$\text{State}$}{
        $f_{\text{layer}}^{(l)} \leftarrow 1.0$
    }\Else{  
        $f_{\text{layer}}^{(l)} \leftarrow 0.0$
    }

    \textbf{Return \(f_{\text{layer}}^{(l)}\)}
\end{algorithm}

\subsubsection{Task-Wise Bias-Guided Strategy}

Although dual-branch collaborative learning can promote feature extraction, our ultimate goal is to improve the generalization performance of segmentation. If the two tasks are treated as completely equal, the reconstruction task may overly dominate the gradient update direction, which can damage segmentation performance. To address this issue, we propose a task-bias guidance strategy as a global regulation mechanism. Specifically, we first compute the Euclidean distances of features before and after model updates along different directions, and use the relative magnitudes of these distances to evaluate feature effectiveness. We define the smaller Euclidean distance as the effective optimization distance of the current model, because it represents the direction toward which the features tend to converge. To clarify the primary and secondary roles of the tasks, we further assign a sign (direction) to this distance according to the optimization objective: if the current optimization trend of the model moves toward the reconstruction model (i.e., is biased to the reconstruction task), we assign a positive value to the effective distance; conversely, if the model tends to deviate from the reconstruction model (i.e., focuses on itself or the segmentation task), we assign a negative value. Based on this logic, we obtain the task interaction factor \(f_{\text{task}}^{(l)}\) for the \(l\)-th layer (see Algorithm 2 for the detailed computation process).

\begin{algorithm}
    \caption{Task-wise Bias-Guided Strategy.}
    \label{alg2}
    \KwIn{training step \(t\); feature maps at intermediate layer \(l\), \(F_{\text{seg}}^{(t,l)}\) and \(F_{\text{rec}}^{(t,l)}\); feature maps after \(k\) update steps, \(F_{\text{seg}}^{(t+k,l)}\) and \(F_{\text{rec}}^{(t+k,l)}\)}

    Compute \(dir_{\text{seg}}\):
    \[
dir_{\text{seg}} \leftarrow 
\begin{cases}
-d\bigl(F_{\text{seg}}^{(t,l)}, F_{\text{rec}}^{(t+k,l)}\bigr), & \text{if } Flag_{\text{seg}} \\
\ \ d\bigl(F_{\text{seg}}^{(t,l)}, F_{\text{seg}}^{(t+k,l)}\bigr), & \text{else}
\end{cases}
\]

    Compute \(dir_{\text{rec}}\):
\[
dir_{\text{rec}} \leftarrow 
\begin{cases}
\ \ d\bigl(F_{\text{rec}}^{(t,l)}, F_{\text{seg}}^{(t+k,l)}\bigr), & \text{if } Flag_{\text{rec}} \\
-d\bigl(F_{\text{rec}}^{(t,l)}, F_{\text{rec}}^{(t+k,l)}\bigr), & \text{else}
\end{cases}
\]

$f_{\text{task}}^{(l)} \leftarrow \text{Sigmoid}_{0-2}\bigl(dir_{\text{seg}} + dir_{\text{rec}}\bigr)$

    \textbf{Return \(f_{\text{task}}^{(l)}\)}
\end{algorithm}

\subsection{Loss Function}
To enhance the dual-branch collaborative mechanism of CoDG-Net and explicitly impose the bias-guided strategy, we propose a novel loss function \(L_{CO}\). This loss computes the discrepancy between intermediate-layer feature maps and adjusts it using the layer-wise guidance factor \(f_{layer}^{(l)}\) and the task-wise guidance factor \(f_{task}^{(l)}\), as follows:

\begin{equation}
    L_{CO} = \frac{1}{N} \sum_{l=1}^{N} f_{layer}^{(l)} \cdot f_{task}^{(l)} \cdot \frac{1}{M} \sum_{i=1}^{M} \left( F_{seg,i}^{(t+k,l)} - F_{rec,i}^{(t+k,l)} \right)^2
\end{equation}

Where \(N\) is the number of intermediate layers, \(M\) is the number of pixels in each feature map, \(f_{layer}^{(l)}\) and \(f_{task}^{(l)}\) are the adjustment factors for the \(l\)-th layer, and \(F_{seg,i}^{(t+k,l)}\) and \(F_{rec,i}^{(t+k,l)}\) denote the \(i\)-th feature value at the \(l\)-th layer of \(S_{seg}\) and \(S_{rec}\) at stage \(t+k\), respectively.

For the loss of \(S_{seg}\), we combine the Dice loss with the collaborative loss \(L_{CO}\) to enhance gradient stability and pixel-level segmentation capability:

\begin{equation}
    L_{seg} = \alpha \cdot L_{Dice} + \beta \cdot L_{CO}
\end{equation}

Where \(\alpha\) and \(\beta\) are the fusion weights of the two losses. Similarly, the total loss of \(S_{rec}\) combines the commonly used MSE loss \(L_{MSE}\) for reconstruction with the collaborative loss \(L_{CO}\):

\begin{equation}
    L_{rec} = \alpha' \cdot L_{MSE} + \beta' \cdot L_{CO}
\end{equation}

Where \(\alpha'\) and \(\beta'\) are the fusion weights of the two losses.

\begin{table*}[htbp]
\centering
\resizebox{\linewidth}{!}{%
\begin{tabular}{lcccccccccccccccc}
\toprule
\multirow{3}{*}{Method} & \multicolumn{8}{c}{Source Domain: T2} & \multicolumn{8}{c}{Source Domain: T1CE} \\
\cmidrule(lr){2-9} \cmidrule(lr){10-17}
& \multicolumn{4}{c}{Dice $\uparrow$} & \multicolumn{4}{c}{HD $\downarrow$} & \multicolumn{4}{c}{Dice $\uparrow$} & \multicolumn{4}{c}{HD $\downarrow$} \\
\cmidrule(lr){2-5} \cmidrule(lr){6-9} \cmidrule(lr){10-13} \cmidrule(lr){14-17}
& Flair & T1 & T1CE & Average & Flair & T1 & T1CE & Average & Flair & T1 & T2 & Average & Flair & T1 & T2 & Average \\
\midrule
No Generalization & 73.37 & 5.02 & 10.69 & 29.69 & 11.96 & 38.65 & 37.44 & 29.35 & 46.64 & 60.26 & 15.67 & 40.85 & 21.56 & 18.03 & 47.76 & 29.12 \\
\cmidrule(lr){1-17}
Fed-DG \cite{liu2021feddg}          & 75.77 & 5.82  & 9.51  & 30.37 & 14.45 & 54.03 & 51.06 & 39.85 & 33.03 & 58.30 & 4.09  & 31.72 & 32.07 & 22.35 & 56.08 & 36.83 \\
SADN \cite{zhou2022generalizable}              & 75.87 & 49.36 & 38.09 & 54.44 & 13.44 & 20.15 & 23.56 & 19.05 & 47.31 & 63.64 & 63.00 & 57.98 & 21.03 & 18.06 & 17.56 & 18.88 \\
EGSDG \cite{jiang2024train}            & 76.16 & 62.43 & 55.87 & 64.82 & 13.43 & 18.84 & 17.79 & 16.69 & 68.49 & \underline{71.22} & \textcolor{red}{\textbf{73.06}} & \underline{70.92} & 17.81 & 12.93 & \underline{14.74} & 15.16 \\
SLAug \cite{su2023rethinking}            & 74.21 & 62.15 & 64.62 & 66.99 & 13.41 & 18.23 & 17.12 & 16.25 & 67.71 & 69.31 & 68.23 & 68.42 & \underline{16.32} & \underline{11.53} & 18.31 & \underline{15.39} \\
MixStyleFlow  \cite{safdari2025mixstyleflow}    & \underline{79.31} & \underline{64.12} & \underline{67.31} & \underline{68.11} & \underline{11.42} & \underline{15.64} & \underline{14.24} & \underline{13.77} & \underline{69.45} & 70.01 & 67.21 & 68.89 & 17.87 & 11.56 & 17.45 & 15.63 \\
\cmidrule(lr){1-17}
CoDG-Net          & \textcolor{red}{\textbf{81.67}} & \textcolor{red}{\textbf{67.43}} & \textcolor{red}{\textbf{68.76}} & \textcolor{red}{\textbf{72.62}} & \textcolor{red}{\textbf{8.23}} & \textcolor{red}{\textbf{12.31}} & \textcolor{red}{\textbf{13.63}} & \textcolor{red}{\textbf{11.39}} & \textcolor{red}{\textbf{72.95}} & \textcolor{red}{\textbf{72.02}} & \underline{71.49} & \textcolor{red}{\textbf{72.15}} & \textcolor{red}{\textbf{10.34}} & \textcolor{red}{\textbf{10.87}} & \textcolor{red}{\textbf{13.97}} & \textcolor{red}{\textbf{11.73}} \\
\bottomrule
\end{tabular}%
}
\caption{Comparative results with other related methods on the BraTS dataset (including using T2 as the source domain and using T1CE as the source domain.  \textcolor{red}{Red} indicates optimal, and \underline{underlining} indicates suboptimal.)}
\label{tab1}
\end{table*}
\begin{table*}[htbp]
\centering
\resizebox{\linewidth}{!}{%
\begin{tabular}{lcccccccccccccccc}
\toprule
\multirow{3}{*}{Method} & \multicolumn{4}{c}{Target Domain: Domain1} & \multicolumn{4}{c}{Target Domain: Domain2} & \multicolumn{4}{c}{Target Domain: Domain3} & \multicolumn{4}{c}{Target Domain: Domain4} \\
\cmidrule(lr){2-5} \cmidrule(lr){6-9} \cmidrule(lr){10-13} \cmidrule(lr){14-17}
& \multicolumn{2}{c}{Disc} & \multicolumn{2}{c}{Cup} & \multicolumn{2}{c}{Disc} & \multicolumn{2}{c}{Cup} & \multicolumn{2}{c}{Disc} & \multicolumn{2}{c}{Cup} & \multicolumn{2}{c}{Disc} & \multicolumn{2}{c}{Cup} \\
\cmidrule(lr){2-3} \cmidrule(lr){4-5} \cmidrule(lr){6-7} \cmidrule(lr){8-9} \cmidrule(lr){10-11} \cmidrule(lr){12-13} \cmidrule(lr){14-15} \cmidrule(lr){16-17}
& Dice & ASD & Dice & ASD & Dice & ASD & Dice & ASD & Dice & ASD & Dice & ASD & Dice & ASD & Dice & ASD \\
\midrule
No Generalization & 82.80 & 19.03 & 64.55 & 26.20 & 85.99 & 19.98 & 76.03 & 20.31 & 88.78 & 16.63 & 84.14 & 10.35 & 85.43 & 9.93 & 68.85 & 25.11 \\ 
\cmidrule(lr){1-17}
RAM-DSIR  \cite{zhou2022generalizable1}        & 95.75 & 7.12  & 85.48 & 16.05 & 89.43 & 13.86 & 78.82 & 14.01 & 94.67 & 7.11  & 87.44 & 9.02  & 94.10 & 7.06  & 85.84 & 8.29  \\
DCAC \cite{hu2022domain}               & 95.54 & 6.35  & 81.43 & 19.20 & 87.85 & 18.28 & 77.72 & 17.15 & 94.28 & 8.11  & 86.80 & 9.14  & \textcolor{red}{\textbf{95.40}} & \underline{5.20}  & 87.68 & 7.12  \\
DFQ  \cite{bi2024learning}              & \underline{96.50} & \textcolor{red}{\textbf{6.01}} & \textcolor{red}{\textbf{87.30}} & 15.72 & 92.52 & 12.09 & 81.92 & 13.05 & \underline{95.04} & \underline{7.05}  & 88.95 & 7.70  & 94.85 & 5.84  & 87.47 & \underline{6.55}  \\
TriD \cite{chen2023treasure}             & 96.39 & 6.43  & 86.49 & \underline{15.65} & 92.56 & 12.04 & 82.23 & 13.09 & 94.77 & 7.10  & 88.97 & 7.68  & 94.20 & 7.01  & 87.62 & 7.13  \\
LangDaug  \cite{tiwary2025langdaug}         & 96.41 & \underline{6.07}  & \underline{87.58} & 15.67 & \underline{92.60} & \underline{11.98} & \underline{82.81} & \underline{13.04} & 95.01 & 7.06  & \underline{89.01} & \underline{7.65}  & 94.33 & 6.98  & \underline{87.71} & 7.10  \\
\cmidrule(lr){1-17}
CoDG-Net           & \textcolor{red}{\textbf{96.53}} & 6.15  & 86.58 & \textcolor{red}{\textbf{15.35}} & \textcolor{red}{\textbf{92.71}} & \textcolor{red}{\textbf{11.87}} & \textcolor{red}{\textbf{83.79}} & \textcolor{red}{\textbf{11.76}} & \textcolor{red}{\textbf{95.11}} & \textcolor{red}{\textbf{6.98}} & \textcolor{red}{\textbf{90.02}} & \textcolor{red}{\textbf{7.38}}  & \underline{95.28} & \textcolor{red}{\textbf{5.17}} & \textcolor{red}{\textbf{88.53}} & \textcolor{red}{\textbf{6.38}} \\
\bottomrule
\end{tabular}%
}
\caption{Comparison of domain generalization results on Optic Disc/Cup segmentation from fundus images (We follow the practice in domain generalization literature to adopt the leave-one-domain-out strategy. \textcolor{red}{Red} indicates optimal, and \underline{underlining} indicates suboptimal.)} 
\label{tab2} 
\end{table*}

\section{Experiments}
\subsection{Experimental Setup}
\noindent
\textbf{Datasets.} We conduct experiments on two medical image domain generalization datasets, covering both single-source and multi-source settings, to verify the effectiveness and generalization ability of our method: the BraTS 2018 dataset \cite{menze2014multimodal} and a fundus image dataset \cite{orlando2020refuge}. The BraTS 2018 dataset contains 210 high-grade glioma cases and 75 low-grade glioma cases. Each case includes four MRI modalities: T2, FLAIR, T1, and T1CE. We use this dataset to perform single-source domain generalization experiments. The fundus image dataset consists of images from four domains collected by different institutions and different scanning devices. The main task is to segment the optic cup and optic disc in fundus images, and this dataset is used for multi-source domain generalization experiments.

\textbf{Implementation details.} All experiments were conducted on Ubuntu 22.04 with Python 3.9 and PyTorch 2.4.1, using 4 NVIDIA RTX 4090 GPUs for training. For a fair comparison, we adopt the same data preprocessing steps as the competing methods. For the BraTS 2018 dataset, we first slice the 3D volumes along the scanning direction, then resize each slice to \(240 \times 240\), and normalize the pixel intensities to the range \([-1, 1]\). The slices are then split into training, validation, and test sets with a ratio of 7:1:2. For the fundus image dataset, we resize the images from \(800 \times 800\) to \(384 \times 384\), and follow the official train–validation split provided by the dataset. For training the diffusion model, we set the number of epochs to 100 and the batch size to 8. We use the Adam optimizer with an initial learning rate of 0.0001, the maximum variance is set to 0.02, and a linear noise schedule is adopted. For training the CoDG-Net collaborative learning network, both the segmentation branch and the reconstruction branch use U-Net as the backbone. The coefficient of \(L_{CO}\) is decayed by a factor of 0.5 every 5 epochs. The random perturbation probability \(\delta\) is set to 0.05, and the hyperparameters \(\alpha, \beta, \alpha', \beta'\) are all set to 0.5.

\textbf{Evaluation metrics.} To evaluate the segmentation performance of the model on target domains, we use the Dice similarity coefficient, Average Surface Distance (ASD), and Hausdorff Distance (HD) for quantitative assessment. To quantitatively measure the severity of catastrophic forgetting, we adopt the concept of the forgetting measure commonly used in Continual Learning (CL) \cite{hayes2018new}. Different from the standard CL setting where tasks arrive sequentially, we define the Source-domain Forgetting Measure (SFM) as the performance gap on the source domain between a baseline model trained only on the source domain (Empirical Risk Minimization, ERM) and the proposed domain generalization model. A lower SFM indicates better preservation of source-domain knowledge. An ideal domain generalization model should significantly improve target-domain performance while keeping SFM close to zero (or even negative, which implies positive transfer).

\subsection{Comparative Experiments}
To comprehensively evaluate the performance of our method, we compare it with existing state-of-the-art (SOTA) approaches on both single-source and multi-source domain generalization tasks, in order to highlight its effectiveness. The experimental results on the BraTS dataset are reported in Table \ref{tab1}, and the results on the Fundus dataset are presented in Table \ref{tab2}.

The results show that our method achieves excellent generalization performance on both single-source and multi-source domain generalization tasks. This is mainly attributed to the proposed structure-guided style augmentation strategy, which can significantly enlarge the coverage of generated samples over unseen domains, enabling the model to better adapt to different domain distributions.

\begin{table}[htbp]
\centering
\resizebox{\linewidth}{!}{%
\begin{tabular}{lcccc} 
\toprule
\multirow{3}{*}{Method} & \multicolumn{2}{c}{Source Domain: T2} & \multicolumn{2}{c}{Source Domain: T1CE} \\
\cmidrule(lr){2-3} \cmidrule(lr){4-5}
& Dice $\uparrow$ & SFM (Dice) $\downarrow$ & Dice $\uparrow$ & SFM (Dice) $\downarrow$ \\
\midrule
No Generalization & 84.52 & -- & 78.71 & -- \\
\cmidrule(lr){1-5} 
Fed-DG           & 68.68 & 15.84 & 59.32 & 19.39 \\
SADN             & 73.71 & 10.81 & 55.15 & 23.56 \\
EGSDG            & 75.93 & 8.59  & 63.67 & 15.04 \\
SLAug            & 73.53 & 10.99 & 60.61 & 18.1  \\
MixStyleFlow     & 75.89 & 8.63  & 62.42 & 16.29 \\
\cmidrule(lr){1-5} 
CoDG-Net         & \textcolor{red}{\textbf{85.29}} & \textcolor{red}{\textbf{-0.77}} & \textcolor{red}{\textbf{78.21}} & \textcolor{red}{\textbf{0.5}} \\
\bottomrule
\end{tabular}%
}
\caption{Comparison of source-domain performance and forgetting rate on the BraTS dataset.}
\label{tab3}
\end{table}

In addition, to further verify the capability of CoDG-Net in alleviating catastrophic forgetting during domain generalization, we compare the SFM values of different methods, as reported in Table \ref{tab3}. The results show that, compared with the baseline model trained only on source-domain data, existing domain generalization methods generally suffer from catastrophic forgetting on the source domain. In contrast, the proposed CoDG-Net can effectively mitigate this problem. 
We also provide a visual comparison of the predictions on the source and target domains in Fig. \ref{result} (results on the Fundus dataset are included in the supplementary material).

\begin{figure}[ht]
  \centering
  \includegraphics[width=\linewidth]{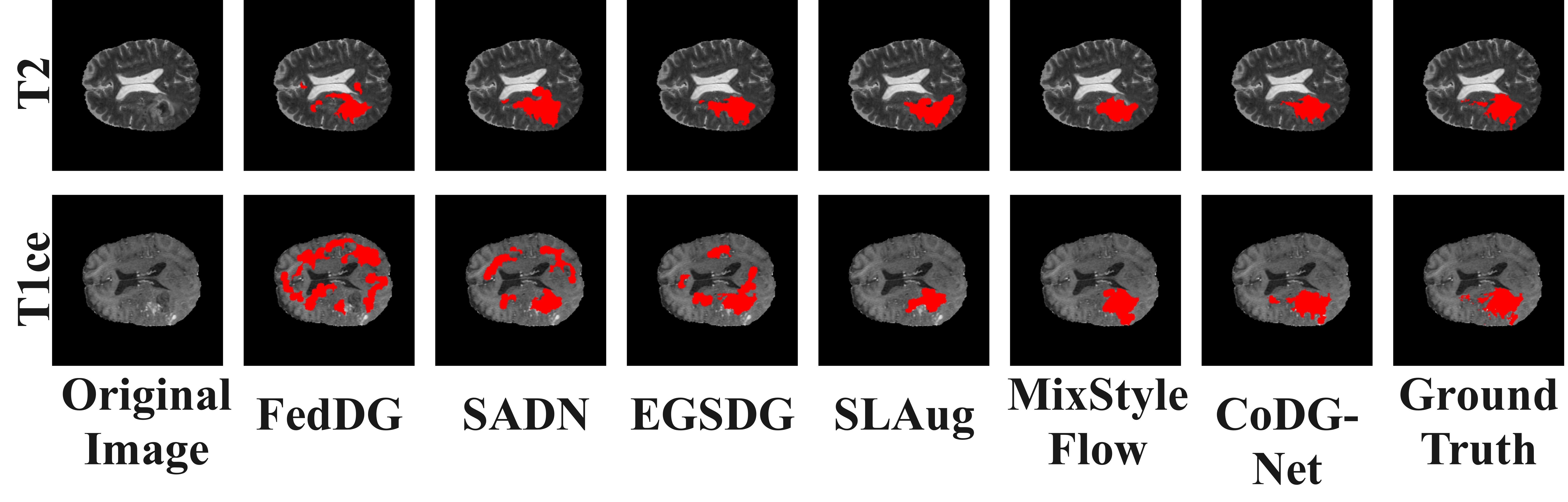}
  \caption{Visual comparison of model predictions between the T2 source domain and the T1ce target domain.}
  \label{result}
\end{figure}

\subsection{Ablation Study and Analysis}
We conduct ablation experiments to evaluate the contribution of the key components in our method, including pseudo-domain style exploration (SE), style diffusion augmentation (SD), dual-branch interaction structure (DBS), layer-wise bias-guided strategy (LW), and task-wise bias-guided strategy (TW). The ablation results are reported in Table \ref{tab4}, where the full model achieves the best performance. Compared with the baseline model (VA1) without domain generalization, introducing the proposed structure-guided style augmentation (VA2, VA3, VA4) significantly improves the performance on the target domain, with Dice increasing from 29.69 to 67.59. However, this also leads to severe forgetting on the source domain, where SFM reaches 5.19. This further validates the rationality of our overall design. It is worth noting that when the SD module is applied only on source-domain data, the degree of source-domain forgetting is relatively small, but the performance gain on the target domain is also limited. This indicates that, without SE, the model can only perform diffusion generation around the source-domain distribution and cannot effectively produce samples from other domains, which demonstrates the necessity of the SE module.

\begin{table}[htbp]
\centering
\resizebox{\linewidth}{!}{%
\begin{tabular}{lccccccccccc}
\toprule
\multirow{2}{*}{Method} & \multirow{2}{*}{SE} & \multirow{2}{*}{SD} & \multirow{2}{*}{DBS} & \multirow{2}{*}{LW} & \multirow{2}{*}{TW} & \multicolumn{2}{c}{T2(Source)} & \multirow{1}{*}{T2-other(Avg)} & \multicolumn{2}{c}{T1CE(Source)} & \multirow{1}{*}{T1CE-other(Avg)} \\
\cmidrule(lr){7-8} \cmidrule(l{0.2em}r{0.2em}){9-9} \cmidrule(lr){10-11} \cmidrule(l{0.2em}r{0.2em}){12-12}
& & & & & & Dice$\uparrow$ & SFM$\downarrow$ & Dice$\uparrow$ & Dice$\uparrow$ & SFM$\downarrow$ & Dice$\uparrow$ \\
\midrule
VA1 & $\times$ & $\times$ & $\times$ & $\times$ & $\times$ & 84.52 & -- & 29.69 & 78.71 & -- & 40.85 \\
VA2 & $\checkmark$ & $\times$ & $\times$ & $\times$ & $\times$ & 80.56 & 3.96 & 51.82 & 74.98 & 3.73 & 44.30 \\
VA3 & $\times$ & $\checkmark$ & $\times$ & $\times$ & $\times$ & 83.16 & 1.36 & 33.35 & 76.42 & 2.29 & 44.13 \\
VA4 & $\checkmark$ & $\checkmark$ & $\times$ & $\times$ & $\times$ & 79.33 & 5.19 & 67.59 & 72.83 & 5.88 & 61.71 \\
\cmidrule(lr){1-12}
VA5 & $\checkmark$ & $\checkmark$ & $\checkmark$ & $\times$ & $\times$ & 81.09 & 3.43 & 52.12 & 75.25 & 3.46 & 56.82 \\
VA6 & $\checkmark$ & $\checkmark$ & $\checkmark$ & $\checkmark$ & $\times$ & 84.40 & 0.12 & 65.99 & 78.19 & 0.52 & 69.39 \\
VA7 & $\checkmark$ & $\checkmark$ & $\checkmark$ & $\times$ & $\checkmark$ & 81.37 & 3.15 & 58.97 & 77.47 & 1.24 & 62.48 \\
\cmidrule(lr){1-12}
VA8 & $\checkmark$ & $\checkmark$ & $\checkmark$ & $\checkmark$ & $\checkmark$ & \textcolor{red}{\textbf{85.29}} & \textcolor{red}{\textbf{-0.77}} & \textcolor{red}{\textbf{72.62}} & \textcolor{red}{\textbf{78.21}} & \textcolor{red}{\textbf{0.50}} & \textcolor{red}{\textbf{72.15}} \\
\bottomrule
\end{tabular}%
}
\caption{Ablation study results on the BraTS dataset.} 
\label{tab4} 
\end{table}

When the dual-branch interaction structure is introduced for collaborative learning, simply enabling interactions at all intermediate layers without any bias guidance (VA5) provides only limited protection of source-domain performance, and even leads to a performance drop on the target domain compared with VA4. This indicates that not all intermediate-layer features possess effective cross-task complementarity; blindly enhancing layer-wise interactions may introduce noisy interference and thus weaken the overall performance of the model. The two bias-guided strategies proposed in CoDG-Net play a crucial role in both preserving source-domain performance and improving the generalization ability of the model. Compared with VA5, VA6 better protects the source-domain performance and avoids introducing additional noise. Compared with VA5, VA7 encourages the model to focus more on the segmentation task, thereby improving its segmentation performance. However, using only one of the two strategies cannot achieve globally optimal performance; only when both strategies are applied simultaneously can the model obtain the best trade-off between source-domain protection and target-domain generalization.

\begin{figure}[ht]
  \centering
  \includegraphics[width=\linewidth]{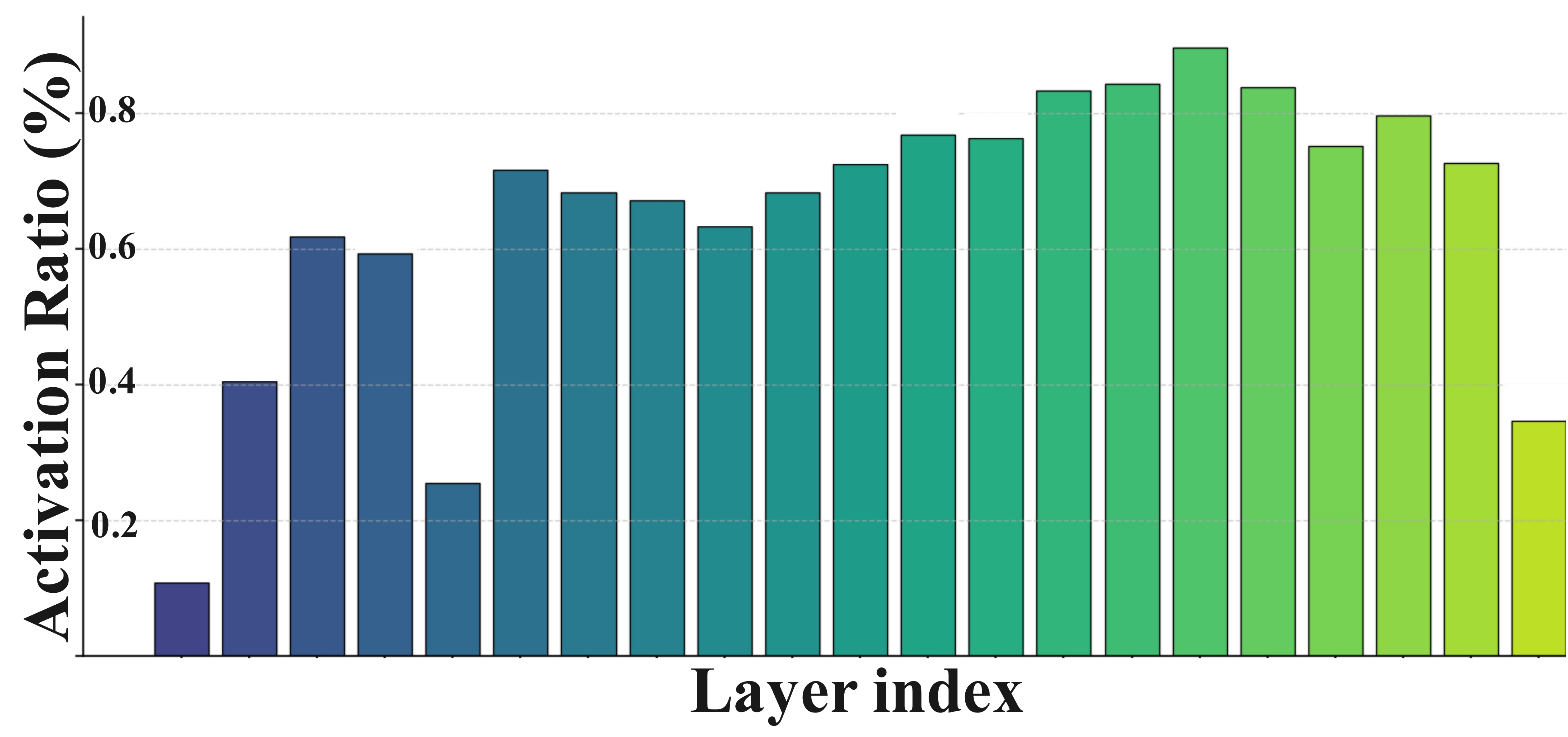}
  \caption{Activation frequency of collaborative interactions at each intermediate layer in CoDG-Net.}
  \label{factor}
\end{figure}

In addition, we analyze the activation frequency of collaborative interactions at each intermediate layer during training, as shown in Fig. \ref{factor}. The vertical axis indicates the proportion of training iterations in which the corresponding layer is in the “open” state. We observe that shallow layers generally have lower activation frequencies, while deeper layers tend to be activated more often, suggesting that not all intermediate-layer interactions are beneficial for segmentation performance. If interactions are blindly and frequently enabled at these layers, it is easy to introduce noisy interference, leading to performance degradation as in VA5. This further demonstrates the necessity of the proposed layer-wise bias-guided strategy.

\begin{figure}[ht]
  \centering
  \includegraphics[width=\linewidth]{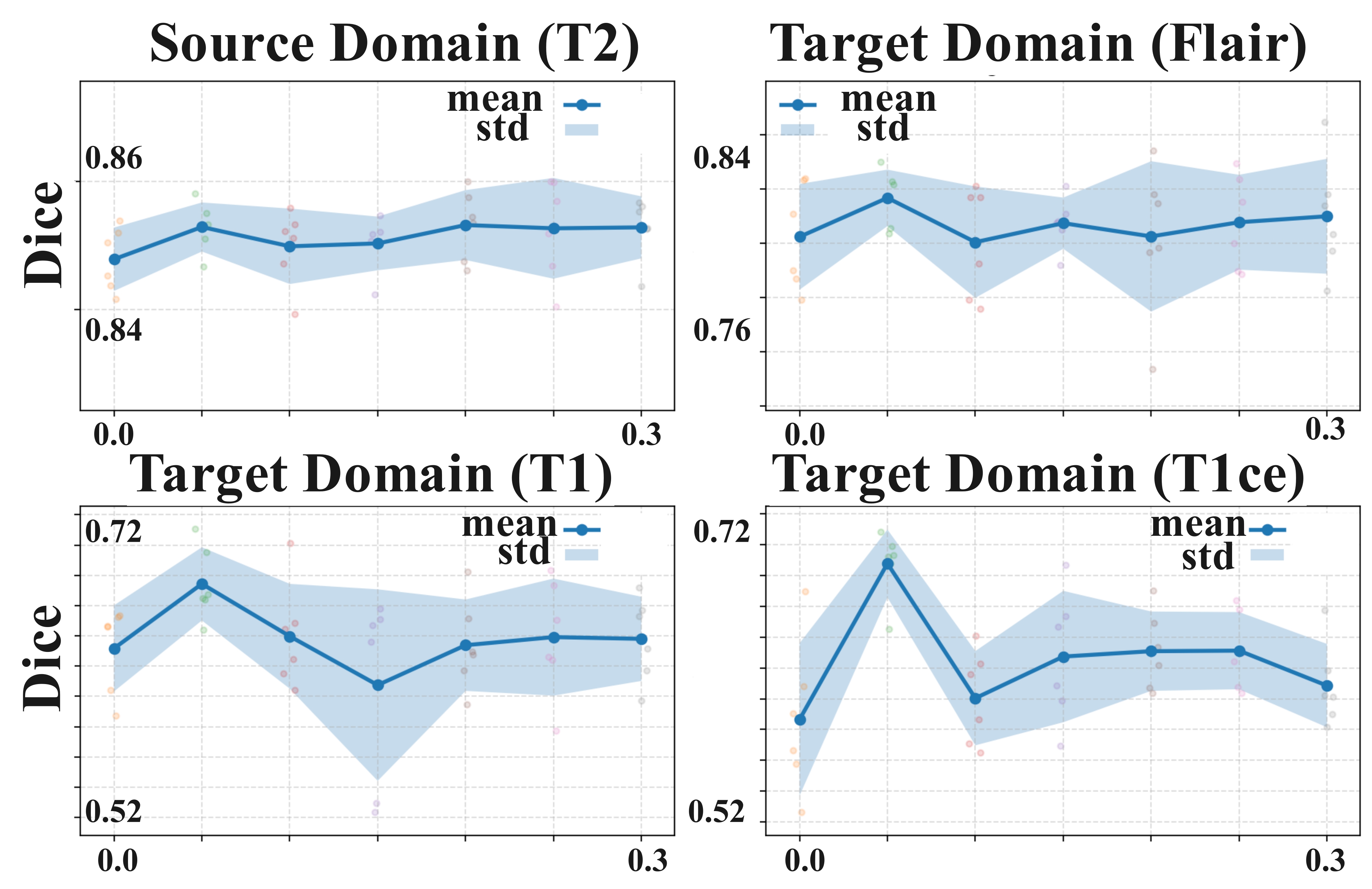}
  \caption{Hyper-parameter sensitivity analysis of the random perturbation probability \(\delta\).}
  \label{hep}
\end{figure}

We also conduct a hyper-parameter sensitivity study on the random perturbation probability \(\delta\), as shown in Fig. \ref{hep}. Using T2 as the source domain on the BraTS dataset, the results show that the T2 source domain is not highly sensitive to \(\delta\). However, when \(\delta = 0.05\), the standard deviation across different random seeds is smaller, indicating more stable model performance. Moreover, on the three target domains (Flair, T1, and T1CE), the model achieves the best performance when \(\delta\) is set to 0.05.

\section{Conclusion}
In this paper, we address the often-overlooked problem of catastrophic forgetting in domain generalization for medical image segmentation and propose a collaborative learning network with a dual-branch interaction structure, termed CoDG-Net. We first introduce a structure-guided style diffusion augmentation method, which provides pseudo-domain samples with more diverse styles and broader coverage for domain generalization. Then, we design a novel bias-guided learning strategy to dynamically regulate knowledge transfer across different layers and between tasks in CoDG-Net, thereby balancing source-domain knowledge retention and segmentation accuracy improvement. Experimental results demonstrate that our method outperforms existing SOTA approaches on both single-source and multi-source domain generalization benchmarks.

\bibliographystyle{named}
\bibliography{ijcai26}

\end{document}